# Max and Coincidence Neurons in Neural Networks

Albert Lee[1] and Kang-L. Wang[1]

[1]University of California, Los Angeles, California, USA

**Abstract**

Network design has been a central topic in machine learning. Large amounts of effort have been devoted towards creating efficient architectures through manual exploration as well as automated neural architecture search. However, today's architectures have yet to consider the diversity of neurons and the existence of neurons with specific processing functions. In this work, we optimize networks containing models of the max and coincidence neurons using neural architecture search, and analyze the structure, operations, and neurons of optimized networks to develop a signal-processing ResNet. The developed network achieves an average of 2% improvement in accuracy and a 25% improvement in network size across a variety of datasets, demonstrating the importance of neuronal functions in creating compact, efficient networks.

## Introduction

Advancements in machine learning have often taken inspiration from biology. Recently, a variety of neurons have been discovered, differing in signal representation, response, structure, as well as transmitters and modulators [1]. Neuronal functions organize into core processing capabilities in biological networks, existing across individual organisms and independent of the host's experience. In contrast, despite various neuron models showing task-specific performance in artificial networks (such as ReLU in image recognition, leaky ReLU in image generation, and gated ReLU in language modeling), neuron models are based predominantly on the integrate-and-fire model. Moreover, the networks today are largely homogeneous, with the same neuron model employed across the entire network.

This research aims to address this gap by developing high-performing networks containing signal-processing neurons. We first create models of *max* and *coincidence* neurons that are compatible with back propagation by approximating the neurons' spiking response to piecewise-linear functions. Afterward, neural architecture search (NAS) is performed to obtain high-performing architectures containing the neurons. To ensure generalizability, we study networks optimized through two bio-inspired NAS algorithms on two hierarchies of datasets. A *growth NAS* continuously increases the network width and depth, allowing us to study networks optimized in an unbounded fashion. A *pruning NAS* progressively removes neurons that contribute least to the network's operation, allowing us to study networks optimized in a bounded, reducing manner. The datasets include visual classification on pictures, hand-written characters, and artwork as well as audio classification on speech, environmental sounds, and music; each with 2 datasets for a total of 12. Finally, we quantitatively analyze top-performing networks to design a signal-processing Resnet (SP-Resnet). The SP-Resnet improves the normalized accuracy over the baseline by an average of 2% (-0.6%~+8%), while simultaneously reducing the number of parameters in the network by 25%.

## Background

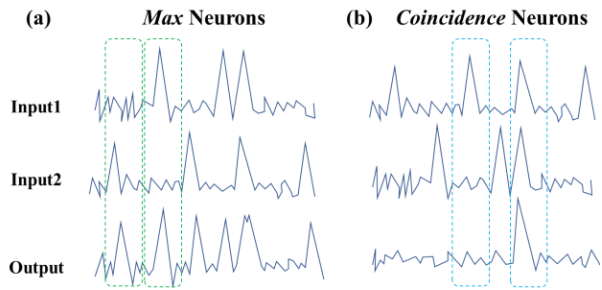

Fig.1 Behavior of the *max* and *coincidence* neurons (a) the *max* neurons output the maximum value among its inputs, and (b) the *coincidence* neurons detect the simultaneous activity among its inputs.

**Max Neurons**

Signal processing in the visual cortex is mainly composed of a hierarchical buildup of receptive fields, in which larger receptive fields of latter layers can be modeled by a linear combination of the smaller receptive fields in earlier layers. However, in the higher regions of the visual cortex, there are neuronal functions that cannot be formed by linear combinations. One such example is the *max* function [2], [3], in which a neuron's output is determined by the input with the highest level of activity (**Fig.1**a). The *max* function can identify the dominant signal within its inputs and eliminate the impact of noise in non-dominant signals, while its commutative property results in a response that is invariant to spatial shifts within its receptive field.

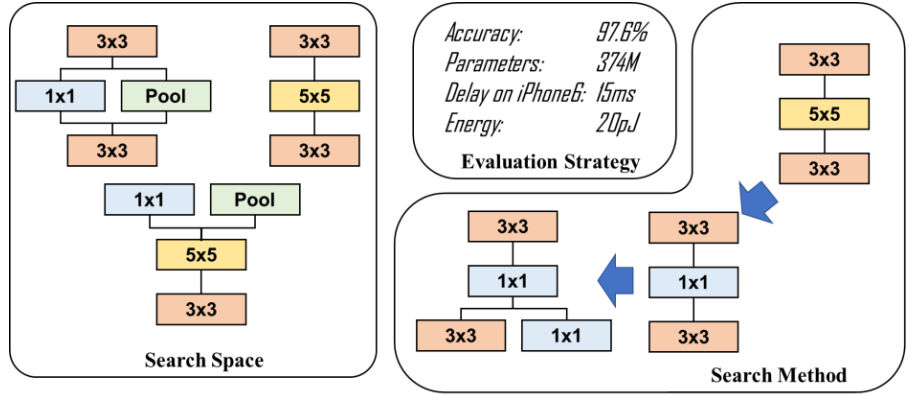

Fig.2 Components of an NAS algorithm, which optimizes a network architecture by iterating through an architecture space. The search space defines the space of architectures the NAS goes through, the search method defines how the architecture changes in each iteration, and the evaluation strategy defines the optimization target of the NAS.

The most successful adaptation of the *max* function today is feature pooling, in which a filter scans localized patches and outputs the maximum value within the region. Feature pooling is usually inserted after convolutions to extract abstract, higher-level concepts from the applied region, and has demonstrated similar benefits as their biological counterparts e.g., dominant signal identification, robustness to noise, and translation invariance [4].

**Coincidence Detection Neurons**

Studies of human auditory processing have shown that temporal and frequency structures are crucial to the representation of audio events [5], [6]. In the human auditory system, cells with extremely low input resistance and time constant detect the *coincident* firing among audio nerve fibers, comparing their temporal correlation with accuracy over 100x of neural firing [7], [8]. They respond maximally when multiple input synapses are simultaneously active and significantly less otherwise (**Fig.1b**).

The success of gating and attention in audio tasks is likely associated with *coincidence* detection [9]–[11]. A gated signal responds maximally when both the gating signal and the input are simultaneously active, while significantly less when either signal is low.

**Neural Architecture Search**

NAS is a technique to optimize network architectures by searching through an architecture space. In general, a NAS algorithm can be described as three components [12]: the search space, the search algorithm, and the evaluation strategy (**Fig.2**). The search space defines the architecture space that the NAS goes through and composes an operation space and a structure space. The former defines the set of operations available to the network, usually including different types of convolutions (2D, separable, depth-wise) and pooling (max, average); with various kernel dimensions (2x2, 3x3, 5x5, 7x7), strides (1,2,4), and dilation rates (1,2,3,4) [13]. The latter defines how the connections between operations are formed, with examples including sequential [14], hierarchical [15], block-based [13], [16], or memory-access-like topologies [17]. Search space design often involves certain amounts of human expertise.

The search algorithm describes how the network architecture changes between search iterations. This could be randomly selecting a network from the search space, a grid search through parameters [18], modification upon a previous top-performing network [15], [19]; or be determined by another network through a reinforcement setting [14], [19]. If the architecture is differentiable, it can also morph via training [20].

The evaluation strategy provides a measure of the architecture's performance. This determines the NAS optimization target and is usually a function of metrics such as accuracy, flops, required memory, energy, latency, etc. Different design of the evaluation strategy allows the NAS to optimize for a single metric or a trade-off between metrics under given constraints. The traditional way to obtain a network's accuracy is to train and evaluate it. As a full train-eval procedure is often costly, accuracy extraction has been accelerated through shortened training schedules on reduced-size networks [14], [16], using pre-trained or shared weights [21], [22], or proxy tasks with lower complexity [14]. Metrics could also be estimated via another network [17], [23].

## Methods

**Neuron Models**

The goal of our modelling algorithm is to create a differentiable model of the neuron that retains its response behavior. This process can be divided into four steps: (1) composing the truth table of the neuron's function, (2) defining numerical values that correspond to the truth tables inputs and outputs (e.g. *spiking/firing*

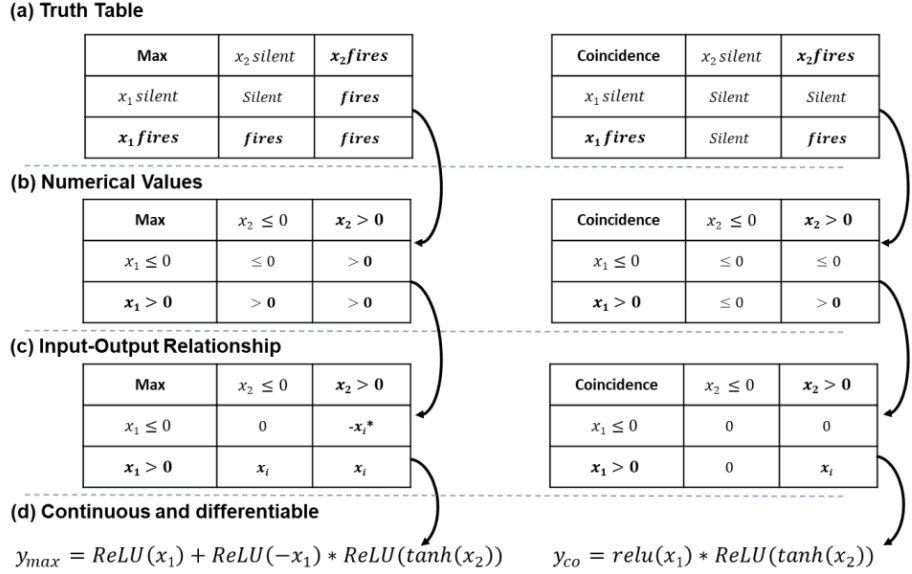

Fig.3 Modelling process of the *max* and *coincidence* neurons, composing four steps: (1) creating the truth table of the neuron's response, (2) determining numerical values of the truth table, (3) determining the input-output relationship of each entry in the table, and (4) making the function differentiable

and *non-spiking/silent*), (3) defining input-output relations for each entry in the truth table, and (4) smoothing discontinuous logic steps. We detail each step below:

Step1: Our first step towards modeling the neurons is to represent their function in the form of a truth table. The truth table will define the regions of linearity and boundaries where nonlinearity occur. Here, we consider a neuron with two inputs $x_1, x_2$ and one output $y$. The *max* neuron's response is the maximum value among its inputs, i.e., its output fires whenever any of its inputs fire. On the other hand, the *coincidence* neurons only fires when all its inputs fire at the same time. The truth tables are shown in **Fig.3a**. Interestingly, the truth tables match that of logical AND and OR, respectively.

Step2. After composing the truth table, we determine the numerical values that correspond to *firing* and *silent*. Here, we follow the conventional definition of *firing* as the signal exceeding a threshold and *silent* otherwise, as shown in **Fig.3b**. The threshold is set to 0.

Step3. We then define the input-output relationship of each truth table entry. Taking inspiration from ReLU, we set the output be linear with unit slope to the first input $x_1$ when the output is *firing* and 0 when it is *silent*, as shown in **Fig.3c.** e.g.,

$$\left|\frac{\partial y}{\partial x_1}\right| \triangleq 1 \ and \ y \triangleq |x_1| \quad y \gg 0$$

$$\left|\frac{\partial y}{\partial x_1}\right| \triangleq 0 \ and \ y \triangleq 0 \quad y \ll 0$$

Step4. Since the truth table contains discontinuous, non-differentiable transitions, we smooth transitions with *tanh* on the second input $x_2$. This gives the two models as follows:

$$y_{co} = ReLU(x_1) * ReLU(tanh(x_2))$$

$$y_{max} = ReLU(x_1) + ReLU(-x_1) * ReLU(tanh(x_2))$$

Where $y_{max}$ is the response of the *max* neuron and $y_{co}$ is the response of the *coincidence* neuron.

As a neuron's response is characterized by an activation function, we place the *max* and *coincidence* models where activations are placed in a neural network, e.g., after the weighted sum or convolution; or after batch normalization if it is used. We let $x_2$ to be the shifted version of $x_1$ in the same layer: for a fully connected layer, shifted by one point, for a 2D convolution, by one in both the first and second dimensions, respectively.

**Growth NAS**

The search space of our growth NAS is based on the multi-branch architecture in [24]. The search method uses the evolutionary backbone of [16] with path-based modifications between search iterations [24], combined with the differentiable architecture search in [20]. The evolutionary backbone primarily optimizes the network architecture, while the differentiable search finds the optimal neurons in the network. Here, we detail the search space, search algorithm, and evaluation strategy of our growth NAS.

**Growth NAS Search Space**

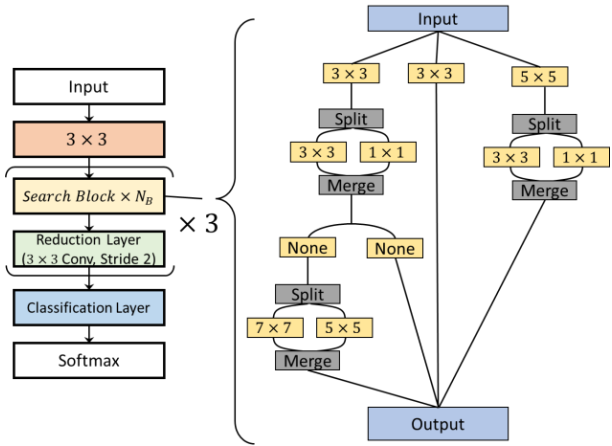

Fig.4 Top level architecture of the growth NAS, e.g., a block-based architecture with *search blocks* generated by *growth* procedures.

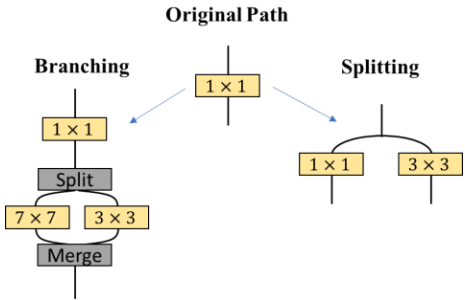

Fig.5 A growth procedure could be *branching* a path or *splitting* a path.

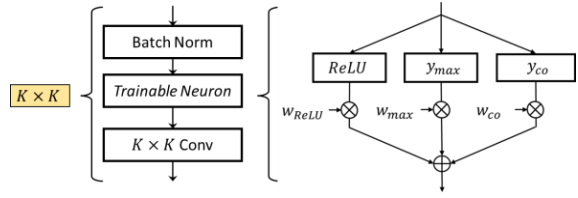

Fig.6 Each operation composes a batch normalization, a *trainable neuron layer*, and a convolution of the corresponding kernel size. *A trainable neuron layer* has sums three weighted paths, each corresponding to a neuron type and scaled by the softmax of the weights

At the top level, the network has a block-based architecture, composing an initial 3x3 convolution followed by three ( $search\ block \times N_B$ – reduction layer) stacks and a classification layer (**Fig.4**). The NAS optimizes the network by manipulating the architecture of the *search block*, each of which has an independent architecture generated through path-level procedures $Branching$ and $Splitting$ (**Fig.5**). $Branching$ a path divides its filters into two groups, applies a random operation to each, then recombines them via concatenation. $Splitting$ a path creates a duplicate of the path with the same input and output connection and a random operation. $Branching$ increases the depth of a path while $Splitting$ increases its width. The available operations include $1 \times 1$ Op, $3 \times 3$ Op, $5 \times 5$ Op, and $7 \times 7$ Op, each composing a convolution of the corresponding kernel size followed by batch normalization and a *trainable neuron layer* (**Fig.6a**), and a *None* Op that passes the input to the output as a shortcut.

The *trainable neuron layer* ( $TL$ , **Fig.6b**) is a trainable layer that finds the optimal neuron type via training. It composes three weighted branches corresponding to the conventional neuron (ReLU), the *max* neuron, and the *coincidence* neuron, respectively. The layer outputs the sum of each branch scaled by the *softmax* of the branch weights ($w_{TL}$), e.g.,

$$y_{TL} = softmax(w_{TL}) \cdot [ReLU(x), y_{max}(x), y_{co}(x)]$$

Every neuron in the trainable neuron layer is individually optimized. For a convolution with $C_{out}$ neurons (or output channels), there are $3C_{out}$ weights.

The reduction layer follows the common implementation of a 3x3 convolution with stride 2, doubled output channels, batch normalization, and a *trainable neuron layer*.

### Growth NAS Search Algorithm

During search, we keep an active population $P$ of $n(P)$ architectures $A_i$ and their evaluation metrics $\epsilon_i$, $i \in [0, n(P)]$. In each search iteration, a sample of size $n(S)$ is sampled from $P$. Within $S$, the architecture $A^*$ with the highest $\epsilon^*$ is selected as the parent. A child architecture $A'$ is then generated by modifying the parent's architecture. The child is then evaluated and added to $P$, and the oldest member of $P$ is removed. Search continues until a maximum number of iterations $n_{iter}$ is met, or when the number of operations in the network exceeds the maximum value $n_{op\_max}$. **Fig.7** shows the pseudocode of the growth NAS.

The initial population $P_{init}$ is composed of random architectures, which are generated by starting with a single $3 \times 3$ Op in each *search block* and applying up to $N_{G\_init}$ *growth* procedures to each block (e.g., $Branching$ or $Splitting$) at random. During each search iteration, $N_{B\_search}$ *search blocks* are randomly selected, and $N_{G\_search}$ random *growth* procedures are applied to each block.

### Growth NAS Evaluation Procedure

Network performances are evaluated a cost-accuracy tradeoff through the evaluation metric $\epsilon = \bar{\alpha}/\bar{\eta}^k$ [25], where $\bar{\alpha} = \alpha_{model}/\alpha_{init}$ is the normalized accuracy of the model, computed as the model accuracy $\alpha_{model}$

**Algorithm1.** Aging Growth NAS

$n_{iter}$: *Number of search iterations*
$n(P)$: *Population Size*
$n(S)$: *Sample Size*

$history.arch \leftarrow [\ ]$
$history.perf \leftarrow [\ ]$
**while** $|history| < n(P)$ **do**
    $model.arch \leftarrow$ **RandomModel**
    $model.perf \leftarrow$ **TrainAndEval**$(model.arch)$
    append $model.arch$ to $history.arch$
    append $model.perf$ to $history.perf$
**end while**
Fit tradeoff hyperparameter $k$ on $history.perf$
**while** $|history| < n_{iter}$ **do**
    $population \leftarrow history[-n(P):]$
    $sample \leftarrow [\ ]$
    **while** $|sample| < n(S)$ **do**
        add **RandomSample**$(population)$ to $sample$
    **end while**
    $parent \leftarrow$ model with highest $perf$ in $sample$
    $child.arch \leftarrow$ **Grow**$(parents.arch)$
    $child.perf \leftarrow$ **TrainAndEval**$(child.arch)$
    append $child.arch$ to $history.arch$
    append $child.perf$ to $history.perf$
**end while**
**Return** highest metric model in $history$

Fig.7 Pseudocode of the growth NAS

**Algorithm 2.** Pruning NAS

$n_{iter}$: *Number of search iterations*

$history.pruned \leftarrow [\ ]$
$history.perf \leftarrow [\ ]$
$model.perf \leftarrow$ **TrainAndEval**$(model.arch)$
append $model.perf$ to $history.perf$
**while** $|history| < n_{iter}$ **do**
    find neuron $N$ with lowest importance metric $\varphi$
    set weights and output of $N$ to 0
    $model.perf \leftarrow$ **FineTuneAndEval**$(model)$
    append $N$ to $history.pruned$
    append $model.perf$ to $history.perf$
**end while**
Return $model$

Fig.9 Pseudocode of the pruning NAS

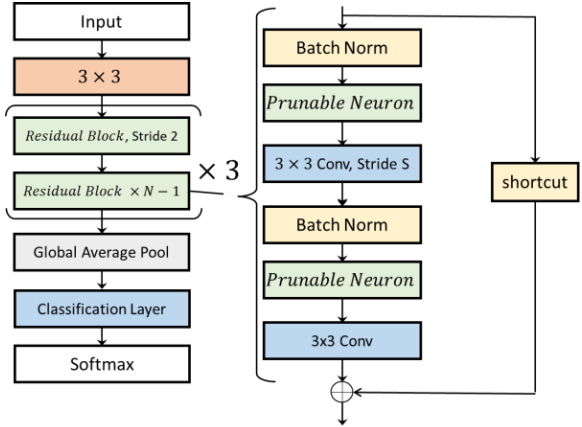

Fig.8 Top level architecture of the pruning NAS, based on a ResNet with an equal amount of each neuron type in the prunable neuron layers.

divided by the average accuracy of the initial population $\alpha_{init}$; $\bar{\eta} = \eta_{model}/\eta_{init}$ is the normalized number of parameters of the model, and $k$ is a hyperparameter for controlling the tradeoff between $\alpha$ and $\eta$, obtained by least-mean-squares fit of $k = -\log \bar{\alpha} / \log \bar{\eta}$ on $P_{init}$.

**Pruning NAS**

The pruning NAS adopts arguably the most widely benchmarked hand-crafted architecture, ResNet [26] as the search space. From a fully trained Resnet with equal amounts of each neuron type, we consecutively remove neurons with low importance based on an *importance metric $\varphi$* until we arrive at an efficient network.

**Pruning NAS Search Space**

The architecture of a Resnet is shown in **Fig.8a**, composing a 3x3 convolution, three ($redisual\ block \times N_B -$ reduction layer) stacks, followed by global average pooling and a classification layer. The *redisual block* architecture is illustrated in **Fig.8b**, composing two $3 \times 3 Ops$ and a shortcut from the block input to the block output. This skip connection provides a shortcut path for signals and gradients by connecting the input and the output. If the block input and block output dimensions match, the input is directly added to the output. Otherwise, a 1x1 convolution is used to map between the block input and output.

Similar to the growth NAS, each $3 \times 3$ Op is built via batch normalization, a *prunable neuron layer (PL)* and a 3x3 convolution. Each *PL* allocates 1/3 of its neurons to each neuron type. During pruning, neurons with the lowest *importance metric $\varphi$* will be removed. The l2 *importance metric* is used e.g., the importance of a neuron is the average of the neuron's squared weights [27].

**Pruning NAS Search Algorithm**

Our pruning algorithm begins with a fully trained network composing an equal amount of each neuron type. In each iteration, we remove $N_R$ neurons with the lowest $\varphi$. Afterwards, the network is fine-tuned via a short training procedure with a reduced learning rate. This process continues until the maximum number of iterations $n_{iter}$ is reached or when the accuracy of the network drops by a predetermined amount $\Delta\alpha_{max}$. The pseudocode of the pruning algorithm is shown in **Fig.9**.

| Dataset | CiFAR-10 | CiFAR-100 | EMNIST | CROHME | Paintings | Wikiart |
|---|---|---|---|---|---|---|
| #Examples ($N_{DS}$) | 50,000 | 50,000 | 700,000 | 300,000 | 7,000 | 64,000 |
| Example Dim.* ($D \times D \times N_c$) | $32 \times 32 \times 3$ | $32 \times 32 \times 3$ | $28 \times 28 \times 3$ | $40 \times 40 \times 3$ | $64 \times 64 \times 3$ | $64 \times 64 \times 3$ |
| $N_F$ | 16 | 16 | 19 | 13 | 6 | 18 |
| $N_S$ | 3900 | 3900 | 14600 | 9600 | 1400 | 4400 |
| Dataset | Commands | Vocalset | ESC | Urbansound | GTZAN | FMA |
| #Examples ($N_{DS}$) | 85,000 | 3,500 | 2,000 | 8,000 | 1,000 | 25,000 |
| Example Dim.* ($D \times D \times N_c$) | $32 \times 32 \times 1$ | $128 \times 128 \times 1$ | $128 \times 128 \times 1$ | $64 \times 64 \times 1$ | $128 \times 128 \times 1$ | $128 \times 128 \times 1$ |
| $N_F$ | 20 | 4 | 4 | 6 | 3 | 12 |
| $N_S$ | 5000 | 1000 | 800 | 1600 | 500 | 2700 |

*After processing

Table I. Dataset characteristics and training settings during the search process

## Experiment Setup

### Data Preparation

The list of datasets used in this study include two categories, each with three subcategories, each with two datasets. For visual recognition, the subcategories are pictures (CiFAR-10, CiFAR-100 [28]), handwritten characters (EMNIST [29], CROHME [30]), and artwork (Best artworks [31], Wikiart [32]). For audio recognition, the subcategories are speech (Speech Commands [33], Vocalset [34]), environmental sounds (ESC [35], Urbansounds8k [36]), and music (GTZAN [37], FMA_medium [38]).

For visual recognition datasets, we adopt the following methods for preprocessing:

1. For datasets without predefined splits (Best artworks and Wikiart), we use a random 80-20 train-eval split.
2. Color images (CiFAR, Best artworks, wikiart) are normalized by their per-channel mean and variance, while greyscale images (EMNIST, CROHME) are standardized to the range [0,1].
3. Since artwork images (Best artworks, Wikiart) vary in dimension, we take five 64x64 crops from each image and treat each as an individual example.
4. No other data augmentation is applied

For audio recognition datasets, we adopt the following methods for preprocessing:

1. For datasets without predefined splits (GTZAN, FMA_Medium, Vocalset), we use a random 80-20 split. For those with predefined folds (ESC, Urbansound8k), we use the first 80% of folds for training and the remainder for evaluation.
2. For each dataset, we clip or pad recordings to the most common length (in seconds) of the dataset. If the recording is over 5 second (Vocalset, ESC, GTXAN, FMA), we divide it into independent 5-second recordings. We then resample the recordings to 16kHz.
3. We compute the log-Mel-Spectorgram of each recording with a number of channels proportional to its length: 1~2sec: 32 channels (Speech Commands), 3~4sec: 64 channels (Urbansounds8k), 5sec: 128 channels (Vocalset, ESC, GTZAN, FMA)). The window size is 512 points and the hop length of 256 points. Afterwards, the sample is linearly interpolated to 32x32, 64x64, or 128x128.

### NAS Parameters

During NAS, it is important to adopt a suitable model size and training setting to avoid over/underfitting. When networks overfit, their performance becomes capped with random fluctuation between different runs, independent of the network architecture. When networks underfit, the non-converged, large variation in accuracy also it difficult to compare architectures. Both overfitting and underfitting would misguide the search process.

In the remainder of this section, we detail the NAS search algorithm settings (i.e., $N_{G\_init}$, $N_{G\_search}$, $N_{B\_search}$, $n(P)$, $n(S)$, $n_{iter}$ etc.), network settings ($N_B$, number of filters $N_F$), and the training/evaluation settings (learning rate schedule, loss functions, number of epochs, etc.) during the search process.

### Growth NAS

Finding optimal NAS settings for every dataset would be prohibitively time consuming. Instead, we extrapolate settings for each dataset from the CiFAR-10 NAS using theoretical studies on how network and training settings should scale with dataset complexity.

In a similar evaluation-based backbone [16], the authors experimented with a variety of population sizes $n(P)$ and sample sizes $n(S)$ and achieved best results via $n(P) = 100$ and $n(S) = 25$; using a network size of $N_F = 24$ and $N_B = 3$. The improvement from NAS was observed to plateau at $n_{iter} = 2 \times 10^4$ iterations. We keep the same $n(P)$, $n(S)$, and $N_B$; but reduce the $N_F$ to 16 to further avoid overfitting. In addition, since our operation space is constrained to 2D convolutions of different sizes (while the reference also includes depthwise and separable convolutions), $n_{iter}$ is reduced to $5 \times 10^2$. To prevent networks from overgrowing, we set $n_{op\_max} = 100$. Finally, as our goal is to study characteristics of high-performing networks containing the signal processing neurons rather than obtain a global optimized network, we focus on exploitation over exploration: $N_{G\_search}$ and $N_{B\_search}$ are both set to 1 and $N_{G\_init}$ is set to 10.

The training settings are as follows: following [16] and [39], we use stochastic gradient descent with a momentum of 0.9 as the optimizer. The initial learning rate is set to 0.01 and decreased by 80% for every 30% of the total training steps (e.g., reduced to $2 \times 10^{-3}$ at the 30th percentile step, $4 \times 10^{-4}$ at the 60th percentile step, and $8 \times 10^{-5}$ at the 90th percentile step). The batch size is 128 and weight decay is $5 \times 10^{-4}$. Since our network is smaller, we the number of training epochs from 25 to 10. Finally, we scale the *trainable neuron layer* weights $w_{TL}$ by 10× to accelerate neuron-type optimization during search.

Using CiFAR-10 as a baseline, we scale the training settings for each dataset based on the results reported in [40]. Empirically and theoretically, the authors suggested that network size and number of training steps is roughly proportional to the square root of dataset size. In addition, we also consider the increase in input complexity that comes from the number of input channels:

$$N_{F,DS} = N_{F,C}\sqrt{(N_{DS}/N_C)}\frac{C_{DS}}{C_C}$$

$$N_{S,DS} = N_{S,C}\sqrt{(N_{DS}/N_C)}$$

Where for a given dataset ($DS$), $N_{DS}$ is the number of training examples, $C_{DS}$ is the number of channels of each example, $N_{F,DS}$ is the network's number of filters and $N_{S,DS}$ is the number of training steps. The reference dataset, CiFAR-10, is denoted with subscript $C$, e.g., $N_C$, $C_C$, $N_{F,C}$, and $N_{S,C}$. The training settings for each dataset is summarized in **Table 1**.

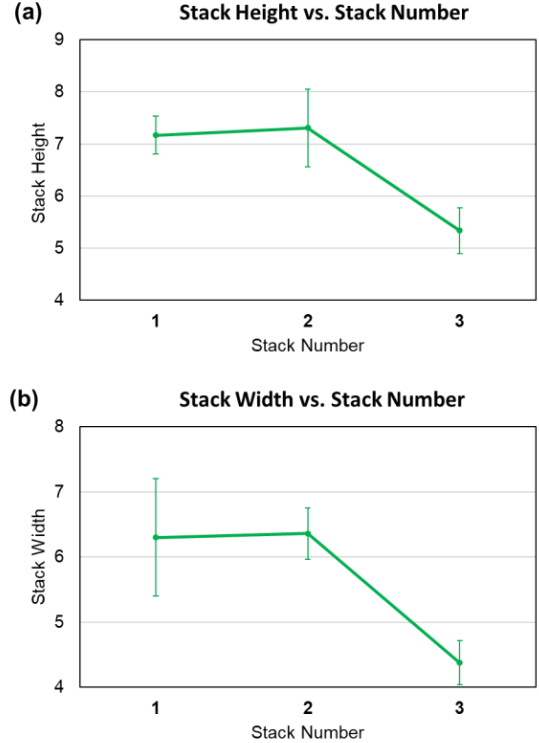

Fig.10 Structural Organization of the optimized networks, (a) height of each stack and (b) width of each stack

**Pruning NAS:**

Similarly, the initial network of the pruning NAS has size of $N_B = 3$ and $N_F = 16 \times 3$ (e.g., $N_F = 16$ for each neuron type). It is trained using the same settings as the growth NAS, but with 50× the number of steps, e.g., 500 epochs, stochastic gradient descent with a momentum of 0.9, initial learning rate of 0.01 decreased by 80% at every 30% of the total number of training steps, batch size of 128 and weight decay of $5 \times 10^{-4}$. The fine-tuning process composes two epochs, one with the decayed learning rate ($2 \times 10^{-3}$) and the other with the double-decayed learning rate ($4 \times 10^{-4}$).

During search, the number of neurons pruned is $N_R = 1$. The maximum number of search iterations $n_{iter} = 1 \times 10^3$, and the maximum recognition rate drop $\Delta\alpha_{max} = 2\%$.

## Results

### Network Statistics

We analyze three characteristics of top-performing networks: (1) *the structural organization* of the architecture, e.g., the height (or depth) and width of each stack as shown in **Fig.4** and **Fig.7**, (2) the *structural composition* of the architecture, e.g., the type and number of operations in each stack, and (3) the *neural composition* of the architecture, e.g., the amount of each type of neurons in each stack. We study the *structural organization* and *structural composition* on networks optimized via the growth NAS, as they have diverse operations and connections; and the *neural composition* via the pruning NAS, which removes the effect of different operations and connections.

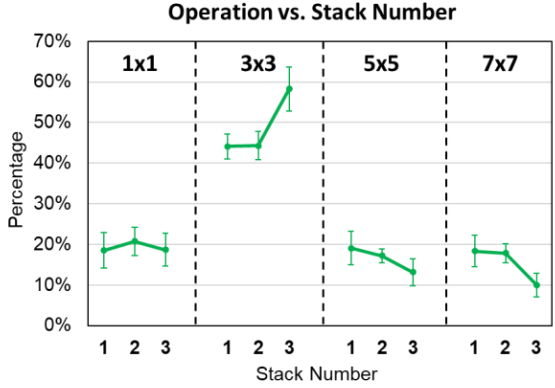

Fig.11 Structural Composition of the optimized networks

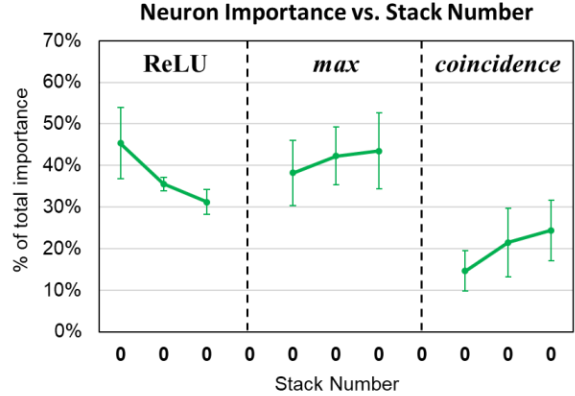

Fig.12 Importance of each neuron type in each stack as a percentage of the total importance of all neurons

**Fig.10** shows the structural organization of top-performing networks found by the growth NAS. Here, we report statistics for each stack (e.g., every $N_B$ search blocks). The height $H$ of an architecture is defined as the longest path from its input to output, and the width $W$ is the maximum number of filters/neurons of the same height normalized to $N_F$. For example, the structure in **Fig.4** has a $H$ of 4 via the leftmost path, and a width of 3 (e.g., at $H = 1$, the operations $3 \times 3$, $3 \times 3$, and $5 \times 5$ have a total to $3N_F$ neurons).

We observe that the average $H$ and $W$ increase slightly from the 1st to 2nd block, but with high variation. Both $H$ and $W$ drop drastically in the 3rd block. This is due the cost tradeoff: the cost (e.g., η) of an operation is proportional to $n_{in} \times n_{out} \times K^2$, where $n_{in}$ and $n_{out}$ are the number of input and output filters, and $K$ is the kernel size. As a result, the cost of a network can be reduced by either (Design-A) adopting a structure with a lower width, which reduces the average $n_{in}$; (Design-B) avoiding operations in the latter stacks as they have larger $n_{in}$ and $n_{out}$; and (Design-C) adopting operations with smaller $K$. Both Design-A and Design-B contribute to the trend observed in **Fig.10**.

**Fig.11** shows the structural composition of the top-performing networks found by the growth NAS. We observe that, in the 1st stack, the $1 \times 1$ Op, $5 \times 5$ Op, and $7 \times 7$ Op occupy a similar percentage while the $3 \times 3$ Op stands out with the highest percentage. Towards the end of the network, the number of operations with small $K$ increases while the number of operations with large $K$ drops significantly, characteristic of Design-B and Design-C: in earlier blocks, having a more diverse $K$ creates an image-pyramid-like representation that can provides features at different scales. However, operations with large $K$ become too expensive in the latter blocks, and thus their application reduces. The $3 \times 3$ Op has the best cost-to-performance tradeoff, which agrees with prior hand-designed networks [38].

**Fig.13** quantifies the neuron importance by showing the total $\varphi$ of each neuron type divided by the total $\varphi$ of all neurons after pruning. We observe that (1) conventional neurons are important in the earlier layers (2) *max* neurons are important in the middle layers, and (3) *coincidence* neurons importance rise in the latter layers.

**Signal-Processing Resnet**

From the analysis results, we propose design modifications to Resnet for a more compact, high-performance architecture. Here, we take the Resnet with $N_B = 3$ and $N_F = 48$ as the baseline and modify the design as described below

(a) Based on the *structural organization* in **Fig.10a**, the network should have an increased 2nd stack height and a decreased 3rd stack height. As a result, we change the number of layers in each stack from [6,6,6] to [6,8,4].
(b) Based on the *structural organization* in **Fig.10b**, the network should have a decreased width in the 3rd stack. As a result, we change the number of filters in the original Resnet from [48,96,192] to [48,96,144].
(c) Based on the *structural composition* in **Fig.11**, we change the operations in the 1st stack from entirely $3 \times 3$ Op to an equal amount of $1 \times 1$ Op, $5 \times 5$ Op, and $7 \times 7$ Op and a slightly higher $3 \times 3$ Op in the 1st stack, e.g., [20%, 40%, 20%, 20%].
(d) Based on the *structural composition* in **Fig.11**, the portion of high kernel size operations

| Dataset | | # Params | CiFAR-10 | CiFAR-100 | EMNIST | CROHME | Paintings | Wikiart |
|---|---|---|---|---|---|---|---|---|
| ResNet ($N_B - N_F$) | 3-32 | 943K | 85.07 | 55.21 | 88.89 | 98.58 | 32.77 | 34.51 |
| | 3-48 | 2.12M | 86.28 | 57.34 | 89.44 | 98.86 | 34.77 | 34.65 |
| | 3-64 | 3.77M | 86.51 | 58.97 | 89.59 | 99.04 | 34.97 | 33.79 |
| | 2-48 | 1.25M | 85.69 | 55.78 | 88.87 | 98.27 | 37.19 | 35.47 |
| | 4-48 | 2.99M | 86.27 | 59.35 | 89.65 | 99.16 | 35.39 | 34.68 |
| ResNet−Struct | | 1.24M | 84.36 | 55.09 | 89.04 | 98.79 | 35.15 | 35.02 |
| ResNet−Op | | 2.11M | 85.54 | 56.83 | 89.08 | 98.79 | 36.47 | 34.58 |
| ResNet−SP | | 1.64M | 86.41 | 59.02 | 89.06 | 98.68 | 35.05 | 34.58 |

| Dataset | | # Params | Speech | Urban | ESC | Vocalset | Gtzan | FMA |
|---|---|---|---|---|---|---|---|---|
| ResNet ($N_B - N_F$) | 3-32 | 943K | 91.22 | 74.54 | 62.23 | 69.43 | 64.32 | 60.73 |
| | 3-48 | 2.12M | 91.26 | 74.41 | 62.76 | 69.18 | 67.40 | 61.35 |
| | 3-64 | 3.77M | 91.69 | 75.52 | 64.32 | 70.47 | 67.52 | 61.59 |
| | 2-48 | 1.25M | 90.99 | 74.41 | 63.80 | 68.60 | 65.62 | 60.81 |
| | 4-48 | 2.99M | 91.05 | 75.91 | 62.50 | 71.11 | 63.01 | 61.43 |
| ResNet−Struct | | 1.24M | 91.26 | 74.93 | 65.10 | 69.24 | 65.64 | 60.30 |
| ResNet−Op | | 2.11M | 91.79 | 73.18 | 63.28 | 71.65 | 66.66 | 61.54 |
| ResNet−SP | | 1.64M | 90.92 | 75.85 | 64.14 | 72.95 | 70.30 | 62.50 |

Table II. Performance of different network designs, including the ResNet of different sizes, the baseline Resnet with only the structural and operation changes, as well as the full signal-processing Resnet

continuously decreases towards the end of the network. We change the percentage of operations in the 2$^{nd}$ and 3$^{rd}$ block to [20%, 50%, 20%,10%] and [30%, 70%, 0%, 0%], respectively.

(e) Based on the *neural composition* in **Fig.12**, we directly apply the percentage of the conventional, *max*, and *coincidence* neurons to be [45%, 40%, 15%], [35%,40%,20%], and [30%,45%,25%] in the 1$^{st}$, 2$^{nd}$, and 3$^{rd}$ blocks respectively.

The performance of different designs is reported in **Table 3** and **Fig.13**. We show the baseline, smaller Resnets with reduced $N_B$ and $N_F$ ($N_B = 2$ and $N_F = 48$; $N_B = 3$ and $N_F = 32$), as well as larger Resnets with increased $N_B$ and $N_F$ ($N_B = 4$ and $N_F = 48$; $N_B = 3$ and $N_F = 64$). To verify the importance of the specialized neurons, we also show the performance of the resnet with only the structural organization modifications ((a) and (b); denoted as $Resnet_{Struct}$), only the structural composition modifications ((c) and (d), denoted as $Resnet_{Op}$), in addition to the full modified Resnet ((a-e) applied, denoted as $Resnet_{SP}$). We observe that without the specialized neurons, neither the change in the structural organization nor composition can outperforms the baseline. On the other hand, when the specialized neurons are employed, we observe an average improvement in normalized accuracy $\bar{\alpha}$ of 2.7% as well as a reduction in number of parameters $\bar{\eta}$ by 25%. In many cases, the proposed network even outperforms networks with increased $N_B$ and $N_F$ that have nearly 2x the number of parameters. These results show the importance of specialized neurons in highly efficient signal processing.

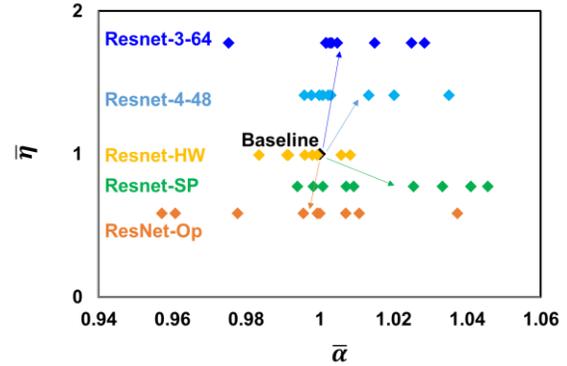

Fig.13 Scatterplot of the performance of each network design

## Summary

The majority of today's networks have yet to consider the diversity of neurons and the importance of specific neuronal responses. In this work, we optimized networks containing signal processing neurons through bio-inspired NAS (Growth, pruning) on a range of datasets, then studied the characteristics of the networks to develop a signal processing Resnet. The developed network demonstrated superior efficiency and outperformed Resnets with ~2x the size. These results show the advantage of heterogeneous networks composed of signal processing neurons, and pave the path towards efficient networks with improved dataflow. One challenge towards these networks be the difficulty of developing accelerators for such heterogeneous networks, which often takes longer time to run.